\documentclass[12pt,a4paper]{article}

\usepackage[english]{babel}

\usepackage[a4paper,top=1in,bottom=1in,left=1in,right=1in]{geometry}
\usepackage{sectsty}
\usepackage{fontspec}

\sectionfont{\fontsize{14}{15}\selectfont}
\subsectionfont{\fontsize{12}{15}\selectfont}

\usepackage{amsmath}
\usepackage{graphicx}
\usepackage[colorlinks=true, allcolors=blue]{hyperref}

\title{Team Fusion\@SU \@ BC8 SympTEMIST track: Transformer-based Approach for Symptom Recognition and Linking}

\author{
   Grazhdanski, Georgi\\
  \texttt{ggrazhdans@uni-sofia.bg}
  \and
  Vassileva, Sylvia \\
  \texttt{svasileva@fmi.uni-sofia.bg}
  \and
  Koychev, Ivan \\
  \texttt{koychev@fmi.uni-sofia.bg}
  \and
  Boytcheva, Svetla \\
  \texttt{svetla@uni-sofia.bg}
}

\begin{document}
\Large Team Fusion@SU @ BC8 SympTEMIST track: Transformer-based Approach for Symptom Recognition and Linking

\vspace{5mm}
\normalsize Georgi Grazhdanski$^{1}$, Sylvia Vassileva$^{1,*}$, Ivan Koychev$^{1}$ and Svetla Boytcheva$^{1,2}$
\vspace{3mm}

$^{1}$FMI, Sofia University St. Kliment Ohridski, $^{2}$Ontotext
\vspace{3mm}

$^{*}$Corresponding author: svasileva@fmi.uni-sofia.bg

\section*{Abstract}
\normalsize This paper presents a transformer-based approach to solving the SympTEMIST named entity recognition (NER) and entity linking (EL) tasks. For NER, we fine-tune a RoBERTa-based \cite{liu2019roberta_og} token-level classifier with BiLSTM and CRF layers on an augmented train set.
Entity linking is performed by generating candidates using the cross-lingual SapBERT XLMR-large \cite{liu2021learning}, and calculating cosine similarity against a knowledge base. The choice of knowledge base proved to have the highest impact on model accuracy.    

\section*{Introduction}

SympTEMIST is a shared task and dataset for detecting and normalizing symptoms, signs, and findings in Spanish clinical texts \cite{symptemistoverview}. It consists of three subtasks - named entity recognition (subtask 1), entity linking (subtask 2), and multilingual entity linking (subtask 3) \footnote{\url{https://temu.bsc.es/symptemist/}}. Our team participated in subtasks 1 and 2.
Previous challenges for named entity recognition in Spanish clinical texts have shown that using a Spanish RoBERTa \cite{spanish_roberta_carrino2021biomedical} classifier with a CRF head on top achieves very good results (\cite{almeida2023}, \cite{chizikova}). Further, adding BiLSTM layer also improves the performance and achieves 79.46\% on the procedures challenge (\cite{almeida2023}).
For the task of entity linking, using cosine similarity search with cross-lingual SapBERT \cite{liu2021learning} embeddings is a common approach utilized by the top three teams in the MedProcNER challenge\footnote{\url{https://temu.bsc.es/medprocner/}} (\cite{vicomtech}, \cite{fusion}, \cite{chizikova}).

\section*{Data}
\subsection*{SympTEMIST Dataset} \label{symptemist_dataset}
The SympTEMIST corpus \cite{symptemistoverview} contains 1,000 clinical case reports in Spanish annotated with symptoms mentions and normalized to SNOMED CT codes. The corpus consists of a fully annotated train set, a smaller test set, as well as a gazetteer of SNOMED-CT codes and different aliases. The train set consists of 750 documents, 11,899 sentences, and 343,243 tokens. The test set has 250 documents, 3,986 sentences, and 114,536 tokens. The train set contains 3,484 annotated entities (2,438 unique), with 1,534 unique entity codes. 59 mentions have no SNOMED CT code assigned, the rest have a single corresponding code. There is 1 nested mention. The Spanish SympTEMIST gazetteer contains a total of 164,817 aliases for terms in multiple categories, including findings, disorders, morphologic abnormalities, and others. 

\subsection*{Language Pre-training Dataset}\label{pretraining_dataset}
To evaluate the effect of further pre-training, we compiled an additional dataset consisting of Spanish UMLS synonyms of all terms included in the SympTEMIST gazetteer, for a total of 337,039 aliases.

\subsection*{UMLS Knowledge Base}
We compiled a knowledge base using all aliases in UMLS Spanish SNOMED-CT which correspond to codes found in the SympTEMIST gazetteer, combined with the gazetteer itself and the train set for subtask 2. The knowledge base consists of 289,734 aliases of symptoms.

\section*{Methods}


\subsection*{Subtask 1 - Named Entity Recognition}
Clinical report texts are split into sentences using the SPACCC Sentence Splitter\footnote{\href{https://github.com/PlanTL-GOB-ES/SPACCC_Sentence-Splitter}{SPACCC Sentence Splitter}}, as about 33\% exceed the 512 input token limit of the employed models.
For the NER subtask we perform token classification following the IOB2 annotation scheme \cite{Krishnan2005NamedER}, using a transformer-based model with a linear layer and a conditional random field (CRF) on top of the last transformer layer. We also experiment with adding a two-layer BiLSTM before the linear layer.

In addition to sentence splitting, some modest data augmentation was applied, by replacing entity mentions with synonyms from the Spanish UMLS, for an additional 1672 example sentences.  

\textbf{Classification Model Selection}

We experiment with the following base models for the token classifier - PlanTL-GOB-ES/roberta-base-biomedical-clinical-es(Spanish RoBERTa) \cite{spanish_roberta_carrino2021biomedical} and CLIN-X-ES \cite{clinx_es_Lange_2022}. The first model is a RoBERTa-based language model, trained on a large Spanish biomedical-clinical corpus of more than 1B tokens. Systems based on this model have achieved competitive results on previous Spanish biomedical-clinical tasks. We further pre-train the model on the \hyperref[pretraining_dataset]{Language Pre-training Dataset}.
The second model, CLIN-X-ES, is a cross-lingual language model, based on XLM-RoBERTa (large), pre-trained on the MeSpEN \cite{mespen_villegas2018mepen} dataset, and Spanish clinical documents from the Scielo archive\footnote{\href{https://scielo.org/}{Scielo archive}}. No additional pre-training is done.

\textbf{Language Model Pre-training}

The effect of further pre-training the base transformer model is evaluated using the \hyperref[pretraining_dataset]{Language Pre-training Dataset} and the PlanTL-GOB-ES/roberta-base-biomedical-clinical-es model with the masked language modeling objective for 4 epochs. Hyperparameter values are the same as those used for pre-training RoBERTa (base) in the original RoBERTa paper\cite{liu2019roberta_og}. 

\subsection*{Subtask 2 - Entity Linking}
The entity linking task uses the gold entities provided by the organizers and aims to predict the correct SNOMED-CT code. For this task, all entities corresponding to more than one code (composite), have been removed from the train and test sets. Furthermore, not all entity mentions have a corresponding code in SNOMED-CT and they are marked as NO\_CODE. There are 59 such instances in the train set.

The system performs linking in two steps - first, we try to match the mention to an alias in the knowledge base (KB), and second - we use cosine similarity search on the SapBERT XLMR-Large \cite{liu2021learning} embeddings and retrieve the closest alias from the KB. 

\section*{Experiments and Results}


\subsection*{Subtask 1 - Named Entity Recognition}
We split the train set and used 80\% for training and 20\% for validation. Micro-averaged precision, recall, and F1-score are used as metrics for the NER subtask.

\begin{table*}[!h]
\caption{Subtask 1 results on the validation and test sets.}
\centering
\begin{tabular}[t]
{ p{7.2cm} p{0.8cm} p{0.9cm} p{1cm} p{1cm} p{1cm} p{1cm} }
\hline
\textbf{Model} & \textbf{Val P} & \textbf{Val R} & \textbf{Val F1} & \textbf{Test P} & \textbf{Test R} & \textbf{Test F1} \\
\hline
Aug. Spanish RoBERTa + CRF & \textbf{0.773} & 0.729 & \textbf{0.750} & 0.732	& 0.718 & 0.725 \\
Aug. Spanish RoBERTa + BiLSTM + CRF  & 0.744 & \textbf{0.732} & 0.738 & \textbf{0.739} & \textbf{0.725} & \textbf{0.732} \\
Pre-trained Aug. Spanish RoBERTa + CRF  & 0.749 & 0.721 & 0.735 & 0.715 & 0.720 & 0.718 \\
CLIN-X-ES + CRF & 0.757 & 0.717 & 0.737 & 0.718	& 0.703 & 0.710 \\
Aug. CLIN-X-ES + CRF & 0.722 & 0.704 & 0.713 & 0.724 & 0.699 & 0.712 \\
\hline
\end{tabular}
\label{tab:subtask1-results}
\end{table*}

Table \ref{tab:subtask1-results} presents the results of the different model and fine-tuning scheme combinations. Models based on the Spanish RoBERTa (PlanTL-GOB-ES/roberta-base-biomedical-clinical-es) perform best, likely due to the fact that it is specialized for the Spanish biomedical-clinical domain. The addition of a 2-layer BiLSTM increases both recall and precision, perhaps because of its ability to consider long-term dependencies\cite{almeida2023}.   

\textbf{Effect of Data Augmentation}

After the training dataset is split into sentences, it is augmented by randomly replacing some of the annotated mentions with a synonym from the Spanish UMLS. This results in 1,672 additional example sentences (13,571 in total). Table \ref{tab:subtask1-augmentation-results} compares model performance. Fine-tuning on the augmented train set significantly improves the precision and F1 of the Spanish RoBERTa model, despite the decrease in recall. Interestingly, we observe a performance drop in the augmented CLIN-X-ES model compared to its non-augmented version on the validation set. However, on the test set, the two models are close in terms of F1, with the augmented CLIN-X-ES having a small precision lead.  



\begin{table*}[!h]
\caption{Effect of data augmentation in the NER subtask.}
\centering
\begin{tabular}[t]{ p{6.4cm} p{1cm} p{1cm} p{1cm} p{1cm} p{1cm} p{1cm} }
\hline
\textbf{Model} & \textbf{Val P} & \textbf{Val R} & \textbf{Val F1} & \textbf{Test P} & \textbf{Test R} & \textbf{Test F1} \\
\hline
Spanish RoBERTa + CRF & 0.730 & \textbf{0.746} & 0.738 & - & - & - \\
Augmented Spanish RoBERTa + CRF & \textbf{0.773} & 0.729 & \textbf{0.750} & \textbf{0.732}	& \textbf{0.718} & \textbf{0.720} \\
CLIN-X-ES + CRF & 0.757 & 0.718 & 0.737 & 0.718	& 0.703 & 0.710 \\
Augmented CLIN-X-ES + CRF & 0.722 & 0.704 & 0.713 & 0.724 & 0.699 & 0.712 \\
\hline
\end{tabular}
\label{tab:subtask1-augmentation-results}
\end{table*}

\subsection*{Subtask 2 - Entity Linking}
We perform experiments with various knowledge bases by combining the different resources - gazetteer, train set, and the UMLS synonym dataset. Furthermore, we generate augmentation to the knowledge by adding/removing random characters in the aliases of rare concepts with less than 5 records in the knowledge base. We generate 5 records for each rare concept.
For validation purposes, we have used the full train set and have not included those examples in the knowledge base. Accuracy metric is used to measure the entity linking.

In addition to using the SapBERT XLMR-Large embeddings, we perform one experiment that aims to tackle the problem of long entity mentions. For each code, we determine the final score as a linear combination of its cosine similarities to the full mention, the first 75\% of tokens in the mention, and the last 75\% of tokens in the mention, and we select the code with the highest score. Using the train set, we find the optimal coefficients with grid search for each part to be 0.75, 0.17, 0.08. The sliding window performs about 2\% better than the basic SapBERT XLMR-Large model using the same knowledge base, which suggests that the information needed to find the correct code is more focused in the first part of the mention.

The results from entity linking experiments are shown in table \ref{tab:subtask2-results}. The best model shows 58.9\% accuracy on the test set and has the richest knowledge base which includes additional data from UMLS. Most of the models show close results in the range of 56-58\% accuracy. Due to a bug in the code for generating the knowledge base, the final experiment shows drastically lower results of about 1\%.

\begin{table*}[!h]
\caption{Subtask 2 results on the validation and test sets.}
\centering
\begin{tabular}[t]{ p{7.2cm} p{5cm} p{0.8cm} p{0.8cm} }
\hline
\textbf{Model} &\textbf{Knowledge Base} & \textbf{Val} & \textbf{Test} \\
 &  & \textbf{Acc} & \textbf{Acc}    \\
\hline
SapBERT XLMR-Large & Gazetteer + Train & 0.514 & 0.588 \\
SapBERT XLMR-Large & Gazetteer + Train + Aug. & 0.533 & 0.565 \\
SapBERT XLMR-Large & Gazetteer + Train + UMLS & 0.524 & \textbf{0.589} \\
SapBERT XLMR-Large + Sliding Window & Gazetteer + Train + Aug. & \textbf{0.536} & 0.587 \\
SapBERT XLMR-Large & Gazetteer + Train + UMLS & 0.510 & 0.017 \\
\hline
\end{tabular}
\label{tab:subtask2-results}
\end{table*}

\section*{Conclusion}
We explore transformer-based approaches to solving the SympTEMIST named entity recognition and linking tasks. For NER, systems based on a specialized monolingual model achieve the best results. The addition of a BiLSTM layer after the last transformer layer, and train data augmentation improves performance on the test set. Further pre-training on a UMLS synonyms dataset did not prove beneficial. We employ SapBERT XLMR-Large exclusively for the entity linking subtask. The choice of a knowledge base has the highest impact on system performance - our highest accuracy model combines the SympTEMIST gazetteer, UMLS synonyms, and train set annotations. Augmenting the knowledge base to include slightly modified versions of rare mentions leads to a stable improvement on the validation set, but has no impact on performance on the test set, so this approach could be explored further. Combining different representations of each entity by using a sliding window may be useful for handling long entity mentions.

\section*{Funding}
This work was supported by the European Union-NextGenerationEU, through the National Recovery and Resilience Plan of the Republic of Bulgaria, project No BG-RRP-2.004-0008.

\bibliographystyle{plain}
\bibliography{sample}

\end{document}